# Learning Fashion Compatibility with Bidirectional LSTMs

Xintong Han, Zuxuan Wu
University of Maryland
College Park, MD
{xintong,zxwu}@umiacs.umd.edu

Yu-Gang Jiang
Fudan University
Shanghai, China
ygj@fudan.edu.cn

Larry S. Davis
University of Maryland
College Park, MD
lsd@umiacs.umd.edu

## ABSTRACT

The ubiquity of online fashion shopping demands effective recommendation services for customers. In this paper, we study two types of fashion recommendation: (i) suggesting an item that matches existing components in a set to form a stylish outfit (a collection of fashion items), and (ii) generating an outfit with multimodal (images/text) specifications from a user. To this end, we propose to jointly learn a visual-semantic embedding and the compatibility relationships among fashion items in an end-to-end fashion. More specifically, we consider a fashion outfit to be a sequence (usually from top to bottom and then accessories) and each item in the outfit as a time step. Given the fashion items in an outfit, we train a bidirectional LSTM (Bi-LSTM) model to sequentially predict the next item conditioned on previous ones to learn their compatibility relationships. Further, we learn a visual-semantic space by regressing image features to their semantic representations aiming to inject attribute and category information as a regularization for training the LSTM. The trained network can not only perform the aforementioned recommendations effectively but also predict the compatibility of a given outfit. We conduct extensive experiments on our newly collected Polyvore dataset, and the results provide strong qualitative and quantitative evidence that our framework outperforms alternative methods.

## KEYWORDS

Fashion recommendation, deep learning, bidirectional LSTM, visual compatibility learning

## 1 INTRODUCTION

Fashion plays an increasingly significant role in our society due to its capacity for displaying personality and shaping culture. Recently, the rising demands of online shopping for fashion products motivate techniques that can recommend fashion items effectively in two forms (1) suggesting an item that fits well with an existing set and (2) generating an outfit (a collection of fashion items) given text/image inputs from users. However, these remain challenging problems as they require modeling and inferring the compatibility relationships among different fashion categories that go beyond simply computing visual similarities. Extensive studies have been conducted on automatic fashion analysis in the multimedia community. However, most of them focus on clothing parsing [9, 26], clothing recognition [12], or clothing retrieval [10]. Although, there are a few works that investigated fashion recommendation [6, 8, 10], they either fail to consider the composition of items to form an outfit [10] or only support one of the two recommendation categories discussed above [6, 8]. In addition, it is desirable that recommendations can take multimodal inputs from users. For example, a user can provide keywords like "business", or an image of a business shirt, or a combination of images and text, to generate a collection of fashion items for a business occasion. However, no prior approach supports multimodal inputs for recommendation.

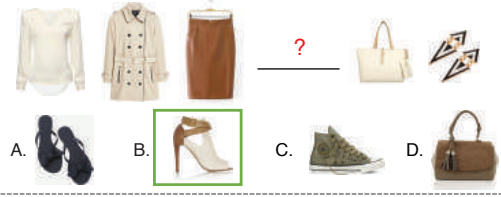

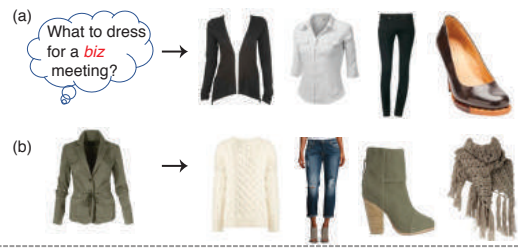

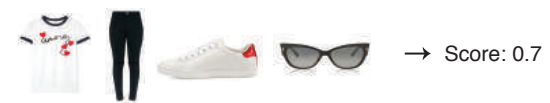

**Figure 1: We focus on three tasks of fashion recommendation. Task 1: recommending a fashion item that matches the style of an existing set. Task 2: generating an outfit based on users' text/image inputs. Task 3: predicting the compatibility of an outfit.**

Key to fashion recommendation is modeling the compatibility of fashion items. We contend that a compatible outfit (as shown in Figure 3) should have two key properties: (1) items in the outfit should be visually compatible and share similar style; (2) these items should form a complete ensemble without redundancy (*e.g.*, an outfit with only a shirt and a pair of jeans but no shoes is not compatible, neither is an outfit containing two pairs of shoes). One possible solution is to utilize semantic attributes [10], for example, "*sweat* pants" matches well with "*running* shoes". But annotating these attributes is costly and unwieldy at scale. To mitigate this issue, researchers have proposed to learn the distance between a pair of fashion items using metric learning [15] or a Siamese network [24]. However, these works estimate pairwise compatibility relationships rather than an outfit as a whole. One could measure the compatibility of an outfit with some voting strategy using all pairs in the set, but this would incur high computational cost when the set is large and would fail to incorporate coherence among all items in the collection. On the other hand, some recent works [8, 21] attempted to predict the popularity or "fashionability" of an outfit, but they fail to handle the outfit generation task. In contrast, we are interested in modeling compatibility relationships of fashion items using their dependencies embedded in the entire outfit.

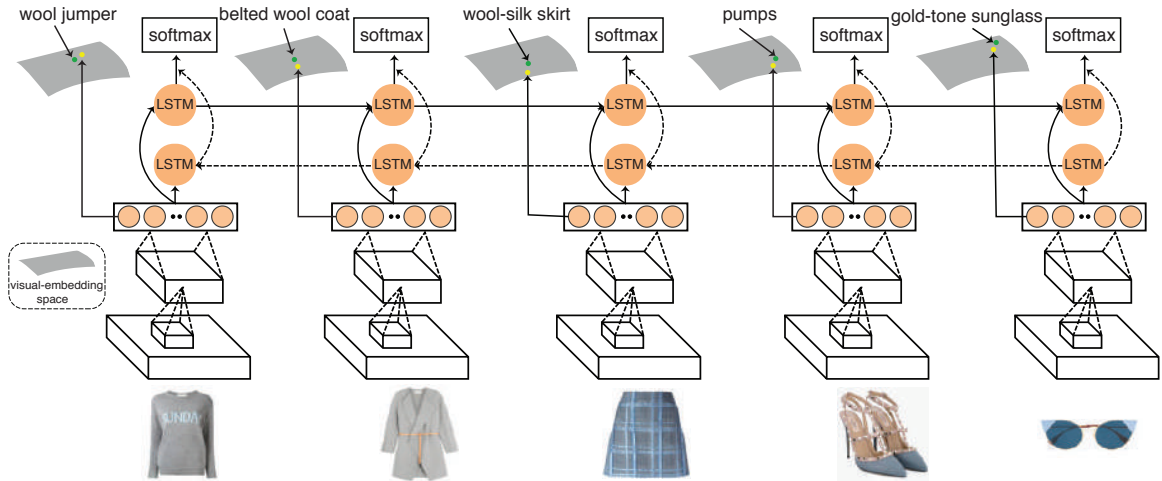

Figure 2: An overview of the proposed framework. We treat a given outfit as a sequence of fashion items (jumper, coat, skirt, pumps, sunglasses). Then we build a bidirectional LSTM (Bi-LSTM) to sequentially predict the next item conditioned on previously seen items in both directions. For example, given the jumper and coat, predict the skirt. Further, a visual-semantic embedding is learned by projecting images and their descriptions into a joint space to incorporate useful attribute and category information, which regularizes the Bi-LSTM and empowers recommendation with multimodal inputs.

To address the above limitations, we propose to jointly learn a visual-semantic embedding and the compatibility relationships among fashion items in an end-to-end framework. Figure 2 gives an overview of the proposed approach. More specifically, we first adopt the Inception-V3 CNN model [22] as the feature extractor to transform an image to a feature vector. Then we utilize a one-layer bidirectional LSTM (Bi-LSTM) with 512 hidden units on top of the CNN model. Bi-LSTM [3] is a variant of Recurrent Neural Networks (RNNs) with memory cells and different functional gates governing information flow, and has have been successfully applied to temporal modeling tasks such as speech recognition [4], and image and video captioning [2, 20]. The intuition of using Bi-LSTM is that we can consider a collection of clothing items as a sequence with a specific order - top to bottom and then on to accessories (*e.g.*, shirt, pants, shoes and sunglasses) - and each image in the collection as a time step. At each time step, given the previous images, we train the Bi-LSTM model to predict the next item in the collection. Learning the transitions between time steps serves as a proxy for identifying the compatibility relationships of fashion items. Furthermore, in addition to predicting the next image, we also learn a visual-semantic embedding by projecting the image features into a semantic representation of their descriptions. This not only provides semantic attribute and category information of the current input as a regularization for training the LSTM, but also enables the generation of an outfit with multimodal inputs from users. Finally, the model is trained end-to-end to jointly learn the compatibility relationships as well as the visual-semantic embedding.

Once the model is trained, we evaluate our network on three tasks as shown in Figure 1: (1) Fill-in-the-blank: given an outfit with one missing item, recommend an item that matches well with the existing set; (2) Outfit generation: generate a fashion outfit with multimodal inputs from the user; (3) Compatibility prediction: predict the compatibility of a given fashion outfit. We conduct experiments on a newly collected Polyvore dataset, and compare with state-of-the-art methods. The main contributions of this work are summarized as follows:

- We jointly learn compatibility relationships among fashion items and a visual-semantic embedding in an end-to-end framework to facilitate effective fashion recommendation in two forms.
- We employ a Bi-LSTM model to learn the compatibility relationships among fashion items by modeling an outfit as a sequence.

- Through an extensive set of experiments, we demonstrate our network outperforms several alternative methods with clear margins.

## 2 RELATED WORK

We discuss multiple streams of works that are closely related to our approach.

**Fashion Recognition and Retrieval**. There is a growing interest in identifying fashion items in images due to the huge potential for commercial applications. Most recent works utilize standard segmentation methods, in combination with human pose information, to parse different garment types [25, 27] for effective retrieval. Liu *et al.* proposed a street-to-shop application that learns a mapping between photos taken by users with product images [11]. Hadi *et al.* further utilized deep learning techniques to learn the similarity between street and shop images [5]. Recently, Liu *et al.* introduced FashionNet to learn fashion representations that jointly predicts clothing attributes and landmarks [12]. In contrast to these works focusing on retrieval tasks, our goal is to learn the visual compatibility relationships of fashion items in an outfit.

**Fashion Recommendation**. As discussed previously, there are a few approaches for recommending fashion items [6, 8, 10]. Liu *et al.* introduced an occasion-based fashion recommendation system with a latent SVM framework that relies on manually labeled attributes [10]. Hu *et al.* proposed a functional tensor factorization approach to generate an outfit by modeling the interactions between user and fashion items [6]. Recently, Li *et al.* trained an RNN to predict the popularity of a fashion set by fusing text and image features [8]. Then they constructed a recommendation by selecting the item that produces the highest popularity score when inserted into a given set. However, the results were no better than random. In contrast to these approaches, our method learns the compatibility relationships among fashion items together with a visual-semantic embedding, which enables both item and outfit recommendation.

**Visual Compatibility Learning**. In the context of fashion analysis, visual compatibility measures whether clothing items complement one another across visual categories. For example, "sweat pants" are more compatible with "running shoes" than "high-heeled shoes". Simo-Serra *et al.* implicitly learned the compatibility of an outfit by predicting its "fashionability" [21]. McAuley *et al.* learned a distance metric between clothes with CNN features to measure their compatibilities [15]. Veit *et al.* further improved the distance metric learning with an end-to-end trained Siamese network [24]. Recently, Oramas *et al.* mined mid-level elements to model the compatibility of clothes [19]. In this paper, we consider the visual compatibility of an entire outfit – items in a fashion collection are expected to share similar styles, forming a stylish composition. To this end, we leverage a Bi-LSTM model to learn the compatibility relationships for outfits, capturing the dependencies among fashion items.

**Sequential Learning with LSTM**. Compared with traditional RNNs, an LSTM is able to model long-range temporal dependencies across time steps without suffering the "vanishing gradients" effect. This results from the use of a memory cell regulated by different functional gates, which assist the LSTM to learn when to forget previous information and when to memorize new things. LSTM models have been successfully applied to capture temporal dependencies in sequences such as speech [4] and videos [2, 18, 28], *etc.* In this work, we employ an LSTM to capture the compatibility relationships of fashion items by considering an outfit as a sequence from top to bottom and then accessories and images in the collection as individual time steps.

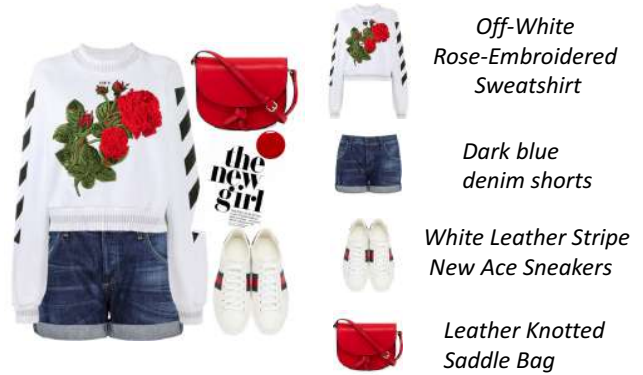

Figure 3: A sample outfit from the Polyvore website. A typical outfit contains a fashion item list, *i.e.*, pairs of fashion images and their corresponding descriptions.

## 3 POLYVORE DATASET

Polyvore (www.polyvore.com) is a popular fashion website, where users create and upload outfit data as shown in Figure 3. These fashion outfits contain rich multimodal information like images and descriptions of fashion items, number of likes of the outfit, hash tags of the outfit, *etc.* Researchers have utilized this information for various fashion tasks [6, 8, 23]. However, their datasets are not publicly available.

Therefore, we collected our own dataset from Ployvore containing 21,889 outfits. These outfits are split into 17,316 for training, 1,497 for validation and 3,076 for testing. Following [8], we also use a graph segmentation algorithm to ensure there are no overlapping items between two splits. For outfits that contain too many fashion items, we only keep the first 8 for simplicity. The resulting Polyvore dataset contains 164,379 items ( each item contains a pair - product image and a corresponding text description). The average number of fashion items in an outfit is 6.5. To clean the text descriptions, we remove words appearing fewer than 30 times, leading to a vocabulary of size 2,757. We choose a large threshold when filtering words because the text descriptions are very noisy and lower-ranked words have very low visualness. Note that the fashion items in an outfit on Polyvore.com are usually organized in fixed order - tops, bottoms, shoes, and the accessories. The orders of the tops and accessories are also fixed - for tops, shirts and t-shirts come before outwears; accessories are usually in the order of handbags, hats, glasses, watches, necklaces, earrings, *etc.* This enables an RNN model like an LSTM to learn "temporal" information. This dataset will be released for research purposes.

## 4 APPROACH

We next introduce the key components of the framework shown in Figure 2, consisting of a bidirectional LSTM for fashion compatibility modeling and a visual-semantic embedding to capture multimodal information.

### 4.1 Fashion Compatibility Learning with Bi-LSTM

The recurrent nature of LSTM models enables them to learn relationships between two time steps, and the use of memory units regulated by different cells facilitates exploiting long-term temporal dependencies. To take advantage of the representation power of LSTM, we treat an outfit as a sequence and each image in the outfit as an individual time step, and employ the LSTM to model the visual compatibility relationships of outfits.

Given a fashion image sequence $F = \{x_1, x_2, ..., x_N\}$, $x_t$ is the feature representation derived from a CNN model for the $t$-th fashion item in the outfit. At each time step, we first use a forward LSTM to predict the next image given previous images; learning the transitions between time steps serves as a proxy for estimating the compatibility relationships among fashion items. More formally, we minimize the following objective function:

$$E_f(F; \Theta_f) = -\frac{1}{N} \sum_{t=1}^{N} \log Pr(x_{t+1}|x_1, ..., x_t; \Theta_f), \quad (1)$$

where $\Theta_f$ denotes the model parameters of the forward prediction model and $Pr(\cdot)$, computed by the LSTM model, is the probability of seeing $x_{t+1}$ conditioned on previous inputs.

More specifically, the LSTM model maps an input sequence $\{x_1, x_2, ..., x_N\}$ to outputs via a sequence of hidden states by computing the following equations recursively from $t = 1$ to $t = N$:

$$i_t = \sigma(W_{xi}x_t + W_{hi}h_{t-1} + W_{ci}c_{t-1} + b_i),$$
$$f_t = \sigma(W_{xf}x_t + W_{hf}h_{t-1} + W_{cf}c_{t-1} + b_f),$$
$$c_t = f_t c_{t-1} + i_t \tanh(W_{xc}x_t + W_{hc}h_{t-1} + b_c),$$
$$o_t = \sigma(W_{xo}x_t + W_{ho}h_{t-1} + W_{co}c_t + b_o),$$
$$h_t = o_t \tanh(c_t),$$

where $x_t, h_t$ are the input and hidden vectors of the $t$-th time step, $i_t, f_t, c_t, o_t$ are the activation vectors of the input gate, forget gate, memory cell and output gate, $W_{\alpha\beta}$ is the weight matrix between vector $\alpha$ and $\beta$ (e.g., $W_{xi}$ is weight matrix from the input $x_t$ to the input gate $i_t$), $b_\alpha$ is the bias term of $\alpha$ and $\sigma$ is the sigmoid function.

Following [16] that utilizes softmax output to predict the next word in a sentence, we append a softmax layer on top of $h_t$ to calculate the probability of the next fashion item conditioned on previously seen items:

$$Pr(x_{t+1}|x_1, ..., x_t; \Theta_f) = \frac{\exp(h_t x_{t+1})}{\sum_{x \in \mathcal{X}} \exp(h_t x)}, \quad (2)$$

where $\mathcal{X}$ contains all images (in multiple outfits) from the current batch. This allows the model to learn discriminative style and compatibility information by looking at a diverse set of samples. Note that one can choose $\mathcal{X}$ to be the whole vocabulary [17] as in sentence generation tasks; however this is not practical during training our model due to the large number of images and high-dimensional image representations. Therefore, we set $\mathcal{X}$ to be all possible choices in the batch of $x_{t+1}$ to speed up training, instead of choosing from hundreds of thousands of images from the training data.

Given a fashion item, it makes intuitive sense that predicting the next item can be performed in the reverse order also. For example, the next item for "pants" could be either "shirts" or "shoes". Therefore, we also build a backward LSTM to predict a previous item given the items after it:

$$E_b(F; \Theta_b) = -\frac{1}{N} \sum_{t=N-1}^{0} \log Pr(x_t|x_N, ..., x_{t+1}; \Theta_b), \quad (3)$$

and

$$Pr(x_t|x_N, ..., x_{t+1}; \Theta_b) = \frac{\exp(\tilde{h}_{t+1} x_t)}{\sum_{x \in \mathcal{X}} \exp(\tilde{h}_{t+1} x)}, \quad (4)$$

where $\tilde{h}_{t+1}$ is the hidden state at time $t + 1$ of the backward LSTM, and $\Theta_b$ denotes the backward prediction model parameters. Note that we add two zero vectors $x_0$ and $x_{N+1}$ in $F$ so that the bidirectional LSTM learns when to stop predicting the next item.

Since an outfit is usually a stylish ensemble of fashion items that share similar styles (e.g., color or texture), by treating an outfit as an ordered sequence, the Bi-LSTM model is trained explicitly to capture compatibility relationships as well as the overall style of the entire outfit (knowledge learned in the memory cell). This makes it a very good fit for fashion recommendation.

### 4.2 Visual-semantic Embedding

Fashion recommendation should naturally be based on multimodal inputs (exemplar images and text describing certain attributes) from users. Therefore, it is important to learn a multimodal embedding space of texts and images. Instead of annotating images with labels or attributes, which is costly, we leverage the weakly-labeled web data, i.e., the informative text description of each image provided by the dataset, to capture multimodal information. To this end, we train a visual-semantic embedding by projecting images and their associated text into a joint space, which is widely used when modeling image-text pairs [7].

Given a fashion image from an outfit, its description is denoted as $S = \{w_1, w_2, ..., w_M\}$ where $w_i$ represents each word in the description. We first represent the $i$-th word $w_i$ with one-hot vector $e_i$, and transform it into the embedding space by $v_i = W_T \cdot e_i$ where $W_T$ represents the word embedding matrix. We then encode the description with bag-of-words $v = \frac{1}{M} \sum_i v_i$. Letting $W_I$ denote the image embedding matrix, we project the image representation $x$ into the embedding space and represent it as $f = W_I \cdot x$.

In the visual-semantic space, we estimate the similarity between an image and its description with their cosine distance: $d(f, v) = f \cdot v$, where $f$ and $v$ are normalized to unit norm. Finally, the images and descriptions are embedded in the joint space by minimizing the following contrastive loss:

$$E_e(\Theta_e) = \sum_f \sum_k \max(0, m - d(f, v) + d(f, v_k)) + \\ \sum_v \sum_k \max(0, m - d(v, f) + d(v, f_k)), \quad (5)$$

where $\Theta_e = \{W_I, W_T\}$ are the model parameters, and $v_k$ denotes non-matching descriptions for image $f$ while $f_k$ are non-matching images for description $v$. By minimizing this loss function, the distance between $f$ and its corresponding description $v$ is forced to be smaller than the distance from unmatched descriptions $v_k$ by some margin $m$. Vice versa for description $v$. During the training, all non-matching pairs inside each mini batch are selected to optimize Eqn. 5. As such, fashion items that share similar semantic attributes and styles will be close in the learned embedding space.

### 4.3 Joint Modeling

Given a fashion output, the Bi-LSTM is trained to predict the next or previous item by utilizing the visual compatibility relationships. However, this is not optimal since it overlooks the semantic information and also prevents users from using multimodal input to generate outfits. Therefore, we propose to jointly learn fashion compatibility and the visual-semantic embedding with an aim to incorporate semantic information in the training process of the Bi-LSTM. The overall objective function is described as follows:

$$\min_{\Theta} \sum_{F} (E_f(F;\Theta_f) + E_b(F;\Theta_b)) + E_e(\Theta_e), \quad (6)$$

where $\Theta = \{\Theta_f, \Theta_b, \Theta_e\}$. The first two terms in Eqn. 6 are the Bi-LSTM objective functions, and the third term computes the visual-semantic embedding loss. The framework can be easily trained by Back-Propagation through time (BPTT) [3] in an end-to-end fashion, in which gradients are aggregated through time. The only difference compared to a standard Bi-LSTM model during back-propagation is that the gradients of the CNN model now stem from the average of two sources (See Figure 2), allowing the CNN model to learn useful semantic information at the same time. The visual-semantic embedding not only serves as a regularization for the training of Bi-LSTM but also enables multimodal fashion recommendation as will be demonstrated in the next section.

## 5 EXPERIMENT

In this section, we first introduce the experiment settings. Then we conduct an extensive set of experiments to validate the effectiveness of the proposed approach on three tasks, including fill-in-the-blank fashion recommendation (Sec. 5.3), fashion compatibility prediction (Sec. 5.4) and fashion outfit generation (Sec. 5.5).

### 5.1 Implementation Details

**Bidirectional LSTM.** We use 2048D CNN features derived from the GoogleNet InceptionV3 model [22] as the image representation, and transform the features into 512D with one fully connected layer before feeding them into the Bi-LSTM. The number of hidden units of the LSTM is 512, and we set the dropout rate to 0.7.
**Visual-semantic Embedding.** The dimension of the joint embedding space is set to 512, and thus $W_I \in \mathbb{R}^{2048 \times 512}$ and $W_T \in \mathbb{R}^{2757 \times 512}$, where 2757 is the size of the vocabulary. We fix the margin $m = 0.2$ in Eqn. 5.
**Joint Training.** The initial learning rate is 0.2 and is decayed by a factor of 2 every 2 epochs. The batch size is set to 10, and thus each mini batch contains 10 fashion outfit sequences, around 65 images and their corresponding descriptions. Finally, we fine-tune all layers of the network pre-trained on ImageNet. We stop the training process when the loss on the validation set stabilizes.

### 5.2 Compared Approaches

To demonstrate the effectiveness of our approach for modeling the compatibility of fashion outfits, we compare with the following alternative methods:
**SiameseNet** [24]. SiameseNet utilizes a Siamese CNN to project two clothing items into a latent space to estimate their compatibility. To compare with SiameseNet, we train a network with the same structure by considering fashion items in the same outfit as positive compatible pairs and items from two different outfits as negative pairs. The compatibility of an outfit is obtained by averaging pairwise compatibility, in the form of cosine distance in the learned embedding, of all pairs in the collection. For fair comparisons, the embedding size is also set to 512. We also normalize the embedding with $\ell_2$ norm before calculating the Siamese loss, and set the margin parameter to 0.8.
**SetRNN** [8]. Given a sequence of fashion images, SetRNN predicts the fashion set popularity using an RNN model. We use the popularity prediction of SetRNN as the set compatibility score.
**Visual-semantic Embedding** (VSE). We only learn a VSE by minimizing $E_e$ in Eqn. 5 without training any LSTM model. The resulting embeddings are used to measure the compatibility of an outfit, similar to SiameseNet.
**Bi-LSTM.** Only a bidirectional LSTM is trained without incorporating any semantic information.
**F-LSTM+VSE.** Jointly training the forward LSTM with visual-semantic embedding, i.e., minimizing $E_f + E_e$.
**B-LSTM+VSE.** Similarly, only a backward LSTM is trained with visual-semantic embedding, i.e. minimizing $E_b + E_e$.
**Bi-LSTM+VSE.** Our full model by jointly learning the bidirectional LSTM and the visual-semantic embedding.

The first two approaches are recent works in this line of research and the remaining methods are used for ablation studies to analyze the contribution of each component in our proposed framework. The hyper-parameters in these methods are chosen using the validation set.

### 5.3 Fill-in-the-blank Fashion Recommendation

Recently, several fill-in-the-blank (FITB) datasets [13, 14, 29, 30] have been created and evaluated to bridge visual and semantic information. However, no existing dataset deals with image sequence completion (i.e., given a sequence of images and a blank, fill in the blank with a suitable image). Thus, in this paper, we introduce the problem of filling-in-the-blank questions from multiple choices as shown in Task 1 of Figure 1. In this task, a sequence of fashion items are provided and one needs to choose an item from multiple choices that is compatible with other items to fill in the blank. This is a very practical scenario in real life, e.g., a user wants to choose a pair of shoes to match his pants and coat.

To this end, we create a fill-in-the-blank dataset using all outfits in the Polyvore test set. For each outfit, we randomly select one item and replace it with a blank, and then select 3 items from other outfits along with the ground truth item to obtain a multiple choice set. We believe that a randomly selected item should be less compatible than the one chosen by experienced designers on Polyvore. Thus, it is reasonable to evaluate fashion recommendation methods on such multiple-choice questions. Once our Bi-LSTM+VSE is trained, we solve the fill-in-the-blank task based on the following objective function:

$$x_a = \arg\max_{x_c \in C} Pr(x_c|x_1, ..., x_{t-1}) + Pr(x_c|x_N, ..., x_{t+1}) \quad (7)$$

$$= \arg\max_{x_c \in C} \frac{exp(h_{t-1}x_c)}{\sum_{x \in C} exp(h_{t-1}x)} + \frac{exp(\tilde{h}_{t+1}x_c)}{\sum_{x \in C} exp(\tilde{h}_{t+1}x)} \quad (8)$$

where $C$ is the choice set, and $t$ is the position of the blank we aim to fill in. Hence, during inference time, forward and backward LSTMs independently predict the probability of one candidate belonging to the outfit, and the candidate having the highest total probability is selected as the answer.

| Method | FITB accuracy | Compatibility AUC |
|---|---|---|
| SetRNN [8] | 29.6% | 0.53 |
| SiameseNet [24] | 52.0% | 0.85 |
| VSE | 29.2% | 0.56 |
| F-LSTM + VSE | 63.7% | 0.89 |
| B-LSTM + VSE | 61.2% | 0.88 |
| Bi-LSTM | 66.7% | 0.89 |
| Bi-RNN + VSE | 63.7% | 0.85 |
| Bi-GRU + VSE | 67.1% | 0.89 |
| Bi-LSTM + VSE (Ours) | **68.6%** | **0.90** |

Table 1: Comparison between our method and other methods on the fill-in-the-blank (FITB) and compatibility prediction tasks.

The middle column of Table 1 shows the results of our method compared with alternative approaches on this task. From this table, we make the following observations: 1) SetRNN and VSE perform similar to random guess (25%); thus they are not suitable for this task. SetRNN predicts popularity of an outfit, but popularity does not always indicate good compatibility. Similar retrieval accuracy is also observed in the SetRNN paper [8]. VSE does not work very well due to the noises in text labels, and also its failure to model the relationships of items in one outfit. 2) SiameseNet works better than VSE and SetRNN but still worse than LSTM based methods, since it mainly considers pairwise relationships rather than the compatibility of the entire outfit; thus it sometimes chooses a candidate with a category that is already in the outfit though the styles are indeed similar. 3) F-LSTM outperforms B-LSTM. We attribute this to the fact that the last several items in most of the outfits are accessories, and it is harder for the backward LSTM to predict clothing items based on accessories than the other way around. The combination of LSTMs in these two directions offers higher accuracy than one directional LSTM. 4) We further jointly learn the Bi-LSTM with the visual-semantic embedding, and the resulting full model achieves the best performance with an accuracy of 68.6%, 1.9 percentage points higher than Bi-LSTM alone. This verifies the assumption the visual-semantic embedding can indeed assist the training of Bi-LSTM by providing semantic clues like classes and attributes. 5) We also investigate different RNN architectures by replacing LSTM cells with gated recurrent unit (GRU) and basic RNN cells. GRU and LSTM are better than basic RNN by better addressing the "vanishing gradients" effect and better modeling the temporal dependencies. The choice between LSTM and GRU depends heavily on the dataset and corresponding task [1]; our experiments demonstrate that LSTM is more suitable for modeling compatibility of fashion items.

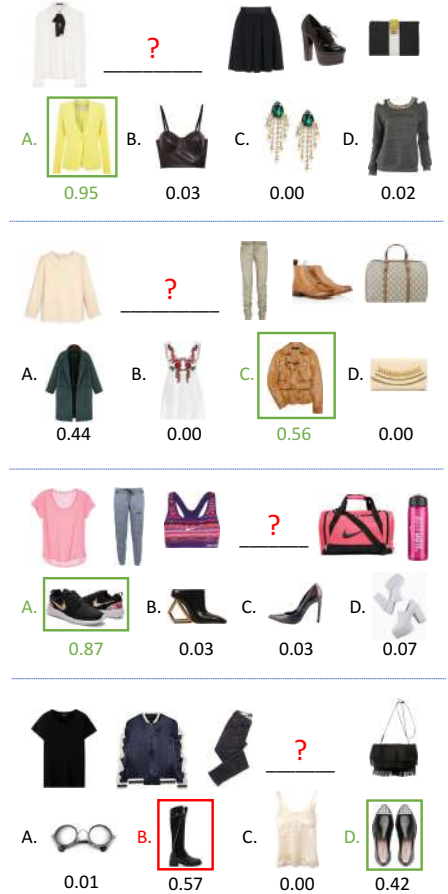

Figure 4: Examples of our method on the fill-in-the-blank task. Green bounding boxes indicate the correct answers, while red box shows a failure case. Prediction score of each choice is also displayed.

In Figure 4, we visualize sample results of our method for the filling-in-the-blank task. Combining Bi-LSTM and visual-semantic embedding can not only detect what kinds of fashion item is missing (*e.g.*, coat is missing in the second example of the Figure 4), but also selects the fashion item that is most compatible to the query items and matches their style as well (*e.g.*, running shoes are more compatible with the sporty outfit in the third example of Figure 4).

### 5.4 Fashion Compatibility Prediction

In addition to recommending fashion items, our model can also predict the compatibility of an outfit. This is useful since users may create their own outfits and wish to determine if they are compatible and trendy. Even though minimizing Eqn. 6 does not explicitly predict compatibility, since our model is trained on the outfit data generated on Polyvore which are usually fashionable and liked by a lot of users, it can be used for this purpose. Given an outfit **F**, we simply utilize the value of the first two terms in Eqn. 6 (Bi-LSTM prediction loss) as an indicator of compatibility.

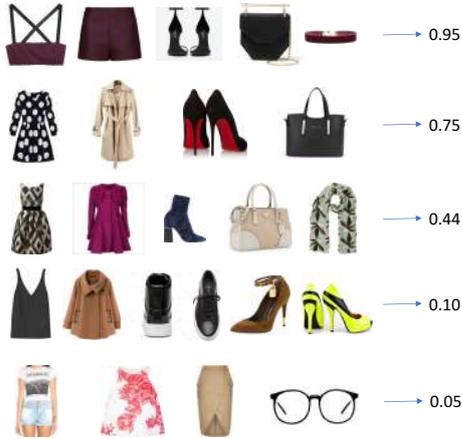

Figure 5: Results of our method on the fashion outfit compatibility prediction task. Scores are normalized to be between 0 and 1 for better visualization.

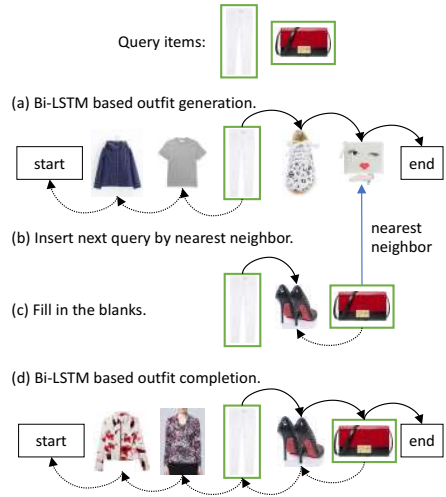

Figure 6: Given query fashion images, our method can generate a compatible outfit.

To compare with alternative methods, similarly to the filling-in-the-blank dataset, we created 4,000 *incompatible* outfits by randomly selecting fashion items from the test set. The performance is evaluated using the AUC of the ROC curve. Results are presented in the third column of Table 1. Our method obtains the best performance among all methods, outperforming recent works [8, 24] by clear margins. Particularly, it is interesting to see that our method, designed to learn the compatibility relationships by predicting the next item conditioned on previous items, is significantly better than SetRNN, which is directly trained to predict set popularity. In addition, we also observe that one directional LSTM is good enough for compatibility prediction.

Figure 5 shows qualitative results of our method. From this figure, we can observe that our method can predict if a set of fashion items forms a compatible (stylish) outfit. For example, the outfit in the first row contains purple/black items with the same style and thus has a high compatibility score; all the items in the third row have different colors, which makes them somewhat incompatible to form an outfit; the fourth outfit contains 4 pairs of shoes without a bottom, and the last one contains two dresses but no shoes; thus they are both incompatible outfits.

### 5.5 Fashion Outfit Generation

We now discuss how to utilize our proposed framework to generate an outfit with multimodal specifications (images/text) from users.
**Generate Outfits from Query Images.** Figure 6 gives an overview of this process. We first consider a degenerate scenario where users provide a single image and wish to obtain an entire outfit with consistent style. This can be accomplished simply by running the trained Bi-LSTM in two directions as shown in Figure 6 (a). When users provide more than one item, we first utilize the first item to generate an initial outfit and then find and replace the nearest neighbor of the next query item (Figure 6 (b)). If the two items are contiguous, we can perform inference in both directions to produce an outfit. Otherwise, we fill in all the blanks between these two items to achieve coherence before performing inference (Figure 6 (c)). This ensures the subsequence used to generate the entire outfit is visually compatible. When more input images are available, this process can be repeated recursively. Finally, the outfit is generated by running the Bi-LSTM model in both directions on the subsequence (Figure 6 (d)). We can see that many fashion items are visually compatible with the white pants, and the initial outfit generated in Figure 6 (a) has a casual style. When incorporating the black/red handbag, our model first predicts a pair of black/red shoes that match both items, and automatically generates an outfit with a slightly more formal style accordingly.

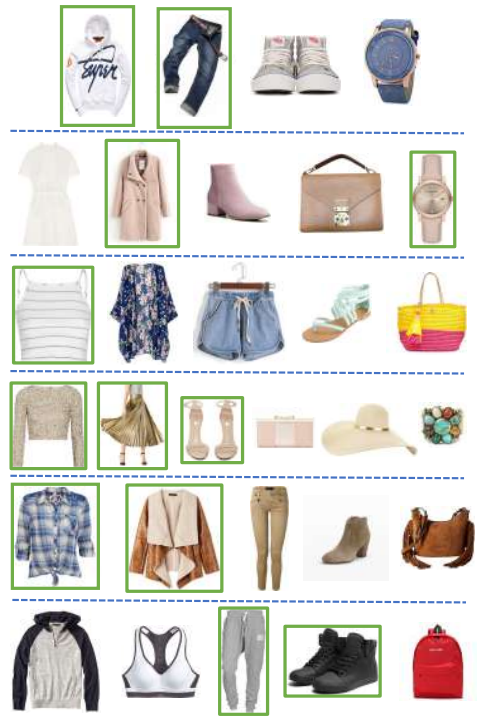

Figure 7: Fashion outfit recommendation given query items. Each row contains a recommended outfit where query images are indicated by green boxes.

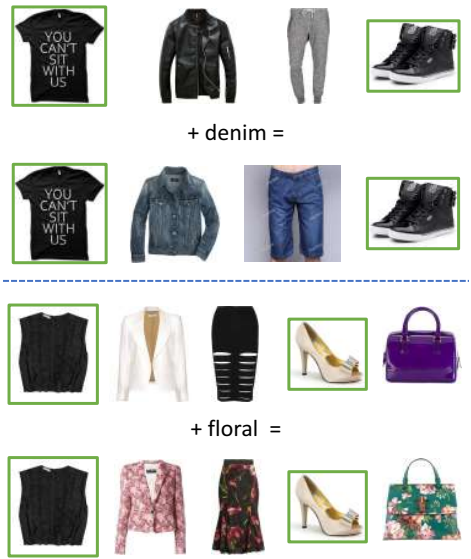

Figure 8: Fashion outfit recommendation given query items and text input. Query images are indicated by green boxes. Outfits on the top are generated without using the text input. When a text query is provided the outfits are adjusted accordingly.

We demonstrate sample outfit generation results given one to three image inputs in Figure 7. It is clear that our method can produce visually compatible and complete outfits. Note that we only show qualitative results of our method since SiameseNet [24], SetRNN [8] and VSE cannot tackle this task.

**Generate Outfits from Multimodal Queries.** Since we jointly learn a visual-semantic embedding together with the Bi-LSTM, our method can also take an auxiliary text query and generate an outfit that is not only visually compatible with the given query fashion items, but also semantically relevant to the given text query. This can be done by first generating an initial outfit using Bi-LSTM based on the given fashion items. Then, given the semantic representation of the text query $v_q$, each non-query item $f_i$ in the initial outfit is updated by $\arg\min_f d(f, f_i + v_q)$. Thus, the updated item is both similar to the original item and also close to the text query in the visual-semantic embedding space. Figure 8 shows two examples of our recommended fashion outfits when multimodal queries are provided. Our model effectively generates visually compatible and semantically relevant outfits.

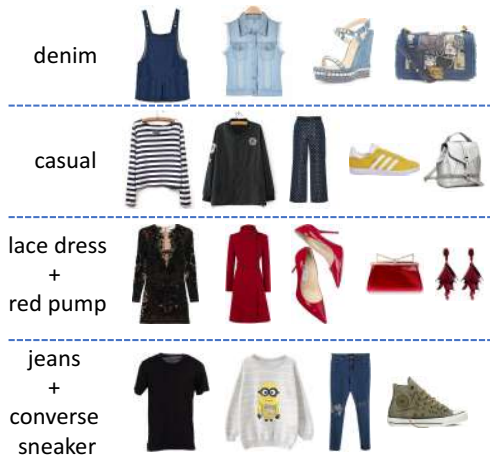

Figure 9: Fashion outfit recommendation given text input. The input can either be an attribute or style (*e.g., denim, casual*) or descriptions of fashion items (*e.g., lace dress + red pump*).

**Generate Outfits from Text Queries.** In addition to combining images and text inputs, our model is also capable of generating outfits given only text inputs. We can take two kinds of text inputs from users - an attribute or style that all items are expected to share, or descriptions of items the generated outfit should contain. In the first scenario, the nearest image to the text query is chosen as the query image, and then the Bi-LSTM model can produce an outfit using this image. Then, the outfit is updated in the same manner as when both image and text inputs are given (the first two examples in Figure 9). In the other scenario, a fashion item image is retrieved using each description, and all images are treated as query images to generate the outfit (the last two examples in Figure 9).

## 6 CONCLUSION

In this paper, we propose to jointly train a Bi-LSTM model and a visual-semantic embedding for fashion compatibility learning. We consider an outfit as a sequence and each item in the outfit as an time step, and we utilize a Bi-LSTM model to predict the next item conditioned on previously seen ones. We also train a visual-semantic embedding to provide category and attribute information in the training process of the Bi-LSTM. We conducted experiments on different types of fashion recommendation tasks using our newly collected Polyvore dataset, and the results demonstrate that our method can effectively learn the compatibility of fashion outfits. Since fashion compatibility might vary from one person to another, modeling user-specific compatibility and style preferences is one of our future research directions.

## ACKNOWLEDGMENTS
The authors acknowledge the Maryland Advanced Research Computing Center (MARCC) for providing computing resources.